\documentclass[preprint]{vgtc}               

\graphicspath{{figures/}{pictures/}{images/}{./}}

\usepackage{times}

\usepackage{mathptmx}
\usepackage{wrapfig}

\newenvironment{myitemize}{
\begin{itemize}
 \setlength{\itemsep}{1pt}
 \setlength{\parskip}{0pt}
 \setlength{\parsep}{0pt}}{\end{itemize}
}

\onlineid{1164}
\vgtccategory{Research}
\vgtcinsertpkg
\preprinttext{}

\title{Toward AI VIS Co-Scientists: A General and End-to-End Agent Harness for Solving Complex Data Visualization Tasks}

\author{Haichao Miao\textsuperscript{1},
        Zhimin Li\textsuperscript{2},
        Kuangshi Ai\textsuperscript{3},
        Kaiyuan Tang\textsuperscript{3},
        Chaoli Wang\textsuperscript{3},
        Peer-Timo Bremer\textsuperscript{1},
        and Shusen Liu\textsuperscript{1}\\ %
        \scriptsize \textsuperscript{1}Lawrence Livermore National Laboratory,
        \textsuperscript{2}Vanderbilt University,
        \textsuperscript{3}University of Notre Dame\\ %
        \scriptsize e-mail: miao1@llnl.gov, zhimin.li@vanderbilt.edu, kai@nd.edu, ktang2@nd.edu, chaoli.wang@nd.edu, bremer5@llnl.gov, liu42@llnl.gov}

\teaser{
  \centering
  \includegraphics[width=\linewidth]{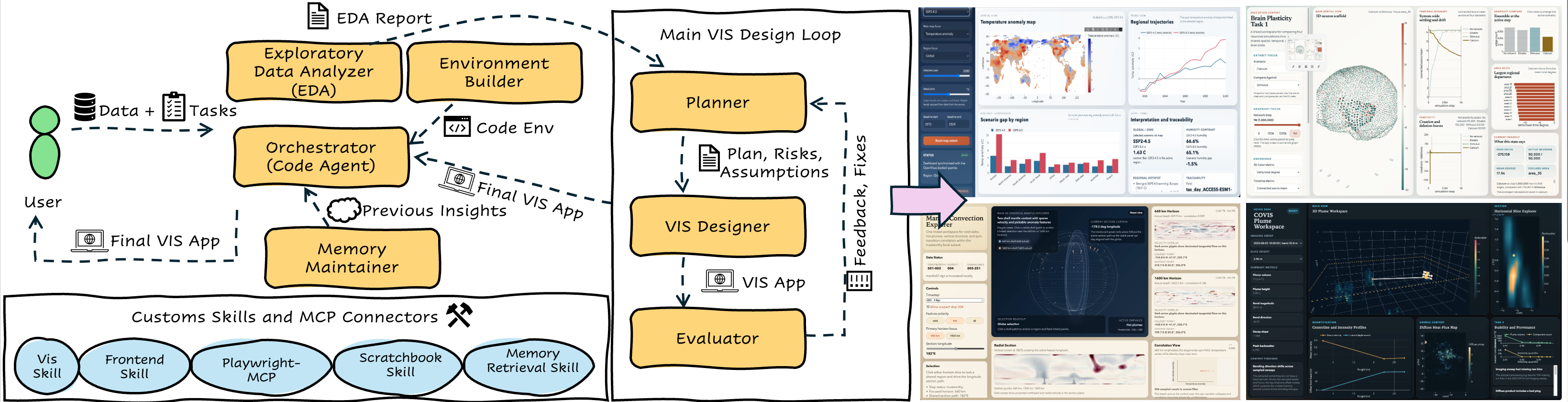}
  \vspace{-0.15in}
  \caption{\textbf{VIS co-scientist agent harness diagram and generated visualization applications (VIS Apps).} Left: Given a dataset and task description, the main code agent orchestrates a multi-stage workflow: specialized subagents, evaluation procedures, and custom skills. The harness enables closed-loop design with browser-based validation. A hierarchical memory system stores insights and lessons learned across sessions. Right: Example outputs from the 2021, 2023, 2024, and 2026 SciVis Contests generated by the VIS co-scientist.}
  \label{fig:teaser}
}

\abstract{
The ability to inspect, interpret, and communicate complex data is crucial for virtually any scientific endeavor, but often requires significant expertise outside the core domain ranging from data management and analysis to visualization design and implementation. We present an end-to-end agentic harness that, based on only the data and a high level description of the tasks, independently designs custom visual analysis applications (VIS apps). This represents an important step towards a general AI co-scientist envisioned by many as an autonomous system that can autonomously execute long horizon tasks based on high-level directions. Our proposed VIS co-scientist is an essential component of this broader AI co-scientist vision: a harness that can autonomously analyze data and design visualization solutions using a collection of agents and specialized skills that coordinate exploratory analysis, plan, configure the environment, implement, validate the interface, and most importantly evaluate the overall task completion. Each stage produces document and instruction artifacts that guide downstream work and enable iterative refinement. We validate this approach on IEEE SciVis Contests spanning multiple science and engineering fields. These contests serve as ideal proving grounds because they encode real-world complexity: ambiguous requirements, diverse data modalities, design trade-offs, and task-driven validation. Given only the data and target tasks, our system autonomously produces functional single-page VIS Apps with verified linked-view behavior, highly customized to domain experts' specified tasks and needs.
}

\keywords{AI agents, scientific visualization, AI scientist, linked-view applications.}

\begin{document}

\firstsection{Introduction}

\maketitle

A variety of disciplines from natural science~\cite{gottweis2025towards, lu2024ai, yamada2025ai, castelvecchi2024researchers, mitchener2025kosmos} to machine learning research~\cite{lu2024ai, yamada2025ai} have started to explore the idea of building AI co-scientists, systems capable of autonomous scientific reasoning, experimentation, and ideally discovery. 
Often these are envisioned as fully automatic systems connecting hypothesis generation to validation and new insights.
However, human exploration, oversight, and input will remain critical not only to bridge remaining gaps in AI capabilities, but to enable validation, provide interpretability, instill trust, and ultimately inspire the next hypothesis. 
Consequently, visual analysis whether in form of plots or complex 3D renderings will remain a crucial component of scientific inquiry. 

In this broader vision of AI co-scientists, we argue that an AI VIS co-scientist is not a niche extension, but an essential component. 
Yet despite growing momentum around AI co-scientists, there has been little concrete effort to explore what such systems could mean for visualization research. Existing AI co-scientist efforts have largely emphasized hypothesis generation and computational research, whereas visualization introduces distinct challenges. Visualization scientists must reason about perceptual principles and design alternatives \cite{li2025perception, esmaeili2022evaluating}, while also translating domain requirements into functional artifacts: data pipelines for large-scale datasets, interactive interfaces for exploratory workflows, and validated applications that genuinely answer scientists' questions. As a result, the VIS co-scientist should be understood as a critical part of the larger AI co-scientist agenda, while also raising unique research questions about which aspects of visualization work can realistically be accelerated by AI and where the major gaps remain.

This work is a first step toward creating the envisioned AI VIS co-scientist by focusing on visual analysis applications (VIS Apps) that represent a common, tangible task performed by most visualization researchers. Despite being a coding-heavy task, current AI coding assistants (e.g., Codex, Claude Code) in their default form are often only suitable for well-defined and isolated subtasks (e.g., benchmark-style tests \cite{ai2026scivisagentbench}). They cannot easily bridge the gap between ``help me understand Atlantic meridional overturning circulation'' and a deployable VIS App with added complexities such as streaming/volumetric data, visual encoding design, coordinated linked views, and interactions. 

To accomplish this goal, we built a specialized agent harness (i.e., the supporting infrastructure and tools around the LLM) in which the main agent orchestrates the workload while delegating focused subtasks to subagents or specialized tools with distinct expertise. An {\em Exploratory Data Analyzer} (EDA) profiles datasets and assesses feasibility. A {\em Planner} translates requirements into concrete specifications. An {\em Environment Builder} configures dependencies. A {\em VIS Designer} handles complex visualization encodings and panel-specific render-debug-verify cycles. An {\em Evaluator} verifies interactive behavior through Playwright-based inspection and grades the VIS App against task goals. The main agent and subagents communicate through explicit artifacts (i.e., reports, plans, instructions, scorecards), making reasoning transparent and enabling iterative refinement.
Additionally, the system explores a prototype hierarchical memory layer inspired by LLM Wiki \cite{karpathy2024llmwiki}. Rather than using complex retrieval strategies, we maintain human-readable markdown files in the form of a wiki that accumulate insights about datasets, tasks, and visualization approaches across sessions. In this short paper, we treat memory primarily as a mechanism for post-run knowledge capture and auditability; measuring whether retrieval improves future VIS design is left for follow-up evaluation. We implement the orchestrating agent using OpenAI Codex, which coordinates the aforementioned specialized subagents and custom \emph{skills} \cite{anthropic2025agentskillsblog}, as well as model context protocol (MCP) \cite{anthropic2024mcp} connectors. 
To evaluate its feasibility, we tested the system on IEEE SciVis Contests from 2021 to 2026, which span climate science, materials discovery, sonar imaging, neuroscience, and mantle convection. These contests demand end-to-end solutions, typically requiring graduate-level skills to address domain-specific tasks. 

Our work represents a clear step toward a VIS co-scientist that can autonomously accomplish long-horizon tasks requiring both complex initial design/planning and iterative refinement.
However, significant challenges remain: the system does not yet exhibit creative visual design beyond established patterns and lacks sophisticated multi-step visual perception and reasoning.
We believe this capability and its trajectory will fundamentally reshape the role of visualization researchers and redefine the frontier where automation ends, and human creativity begins.  
This work serves dual purposes: demonstrating technical feasibility through a structured multi-agent harness and positioning what VIS co-scientists must achieve to become genuine research partners.
Our key contributions include the following:
\begin{myitemize}
  \item \textbf{Introduce VIS co-scientist concept and explore feasibility}: Analysis of what current AI systems achieve and fundamental gaps that remain (i.e., creative design, perceptual optimization), providing a roadmap for future AI VIS co-scientists and characterizing current model shortcomings.
.  \item \textbf{Propose a novel agent harness for end-to-end visualization design}: A re-usable suite of tools and scaffolding that allow a main agent to orchestrate data exploration, planning, environment setup, implementation, browser validation, and evaluation, with artifacts supported by custom skills and MCP connectors.


\end{myitemize}

\section{Related Work}

\paragraph{AI Scientists and AI in Scientific Workflow.} 
AI-based scientific systems~\cite{lyu2026evoscientist,weng2025deepscientist,shao2025omniscientist, gottweis2025towards, lu2024ai, yamada2025ai, castelvecchi2024researchers, mitchener2025kosmos, tang2025ai} have been introduced to accelerate research cycles, including hypothesis refinement, code generation, and knowledge discovery, and to generate research papers that meet top-tier conference standards.  
The concept of AI scientists was first introduced by Lu et al.~\cite{lu2024ai}, which supports tasks such as background research, experiment refinement, and manuscript writing. 
EvoScientist~\cite{lyu2026evoscientist} introduces multi-agent scientists that guide idea direction and validation, and improve experimental strategy.
Most notably, AI co-scientist~\cite{gottweis2025towards} builds on Gemini2 models, which leverage collaboration among multiple agents to improve hypothesis quality and generate experimental insights.

\paragraph{AI Benchmarks for Visualization Tasks.}
In the last few years,  a substantial body of work has been dedicated to the development of AI agents~\cite{liu2024ava,liu2025paraview,ai2025nli4volvis,sun2026sasav,mallick2024chatvis} for visualization exploration and interpretation. Ai et al.~\cite{ai2025evaluation} advocate an evaluation-centric paradigm, arguing that systematic, scalable evaluation is essential to accelerating the development of visualization agents. They further introduce SciVisAgentBench~\cite{ai2026scivisagentbench}, which comprises 108 cases covering various SciVis tasks.
Visualization literacy~\cite{lee2016vlat,pandey2023mini} was originally designed to assess humans’ ability to interpret data visualizations. Recent studies~\cite{bendeck2024empirical,hong2025llms,li2025see} have applied such benchmarks to evaluate LLM models’ capabilities in chart understanding.
SVLAT~\cite{do2026svlat} introduces a framework for SciVis literacy, addressing the limitation that prior literature primarily focuses on the InfoVis domain. VisEval~\cite{chen2024viseval} proposes a benchmark for visualization generation and introduces an automatic evaluation framework.
\section{Building the VIS Co-Scientist}
\label{sec:method}

\begin{wrapfigure}{r}{0.5\linewidth}
    \centering
    \includegraphics[width=\linewidth]{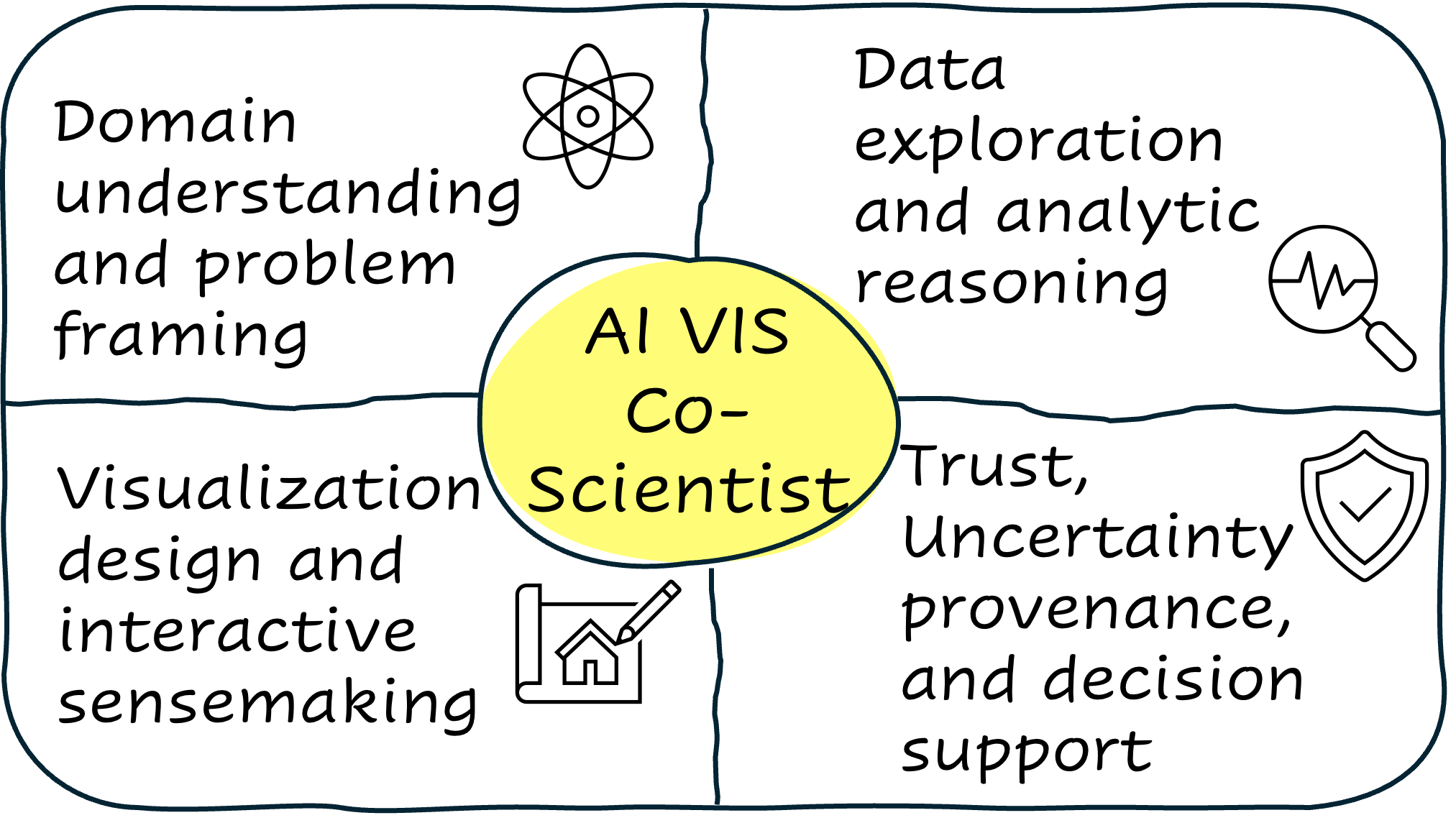}
    \vspace{-6mm}
    \caption{The key capabilities for an AI VIS co-scientist.}
    \label{fig:VIS_co_scientist}
\end{wrapfigure}

Many existing discussions of AI co-scientists focus heavily on hypothesis generation in the context of experimental science \cite{gottweis2025towards}. However, the field of visualization calls for a uniquely broad set of capabilities. Visualization research is inherently multidisciplinary, as its primary goal is to facilitate understanding and discovery in other domains. Compared with many other fields that can rely on quantitative metrics, such as prediction accuracy in machine learning, visualization quality is much harder to measure and evaluate. Moreover, visualization is fundamentally human-centered, requiring effective communication and interaction with a domain expert to support decision-making. These unique requirements can be broadly grouped into four capabilities, summarized in Figure~\ref{fig:VIS_co_scientist}, that help define the scope. 
We define a VIS co-scientist as a research assistant who can independently: 1) obtain domain understanding and translate tasks into concrete requirements; 2) carry out self-directed exploratory data analysis and analytical reasoning; 3) design and implement a complete, fully featured interactive visualization tool; and 4) reliably assess the quality and robustness of the analysis, while effectively communicating data uncertainty to support human decision-making. Each of these capabilities is challenging on its own for current agentic AI systems. Our effort focuses primarily on the first three areas, centered around application-driven visualization system design.

\subsection{Multi-Agent Harness for Visualization and Analysis}

To support the multi-faceted capabilities discussed above, we implement VIS co-scientist as a multi-agent harness: a structured execution environment around a main LLM code agent that defines agent roles, tool access, artifact handoffs, and evaluation loops. Given data and tasks, the harness produces a final VIS App through the stages shown in the illustration (Fig. \ref{fig:teaser}): exploratory data analysis, environment configuration, the main VIS design loop, and evaluation. A prototype memory-maintenance step records lessons after a run, but the reported application results do not yet depend on prior memory retrieval. Each stage emits an artifact that either constrains the next stage or documents evidence used for later verification. This multi-agent setup aligns with Anthropic's recommendation on long-running agent design \cite{anthropic2026longrunning}. Our agent harness is released as open source: \textit{(currently waiting upon code review completion at our organization)}

\paragraph{Orchestrator (Code Agent).}
The Orchestrator owns the end-to-end run. It reads the task materials, launches subagents, integrates their artifacts, and writes most of the application code. Its implementation target is a single-page VIS App with coordinated views, shared interaction state, and a desktop-style layout. The Orchestrator also performs routine fixes directly, while delegating bounded visualization or evaluation problems when specialized attention is useful and to keep its own context window focused.

\paragraph{Exploratory Data Analyzer (EDA).}
The EDA agent profiles the dataset before design begins. It identifies data schema, scale, missing values, file formats, access constraints, and task-relevant variables. Its output is an extensive EDA Report that becomes the empirical basis for later detailed design planning. 

\paragraph{Environment Builder.}
The Environment Builder creates the Code Environment (i.e., dependencies) required by the planned VIS App. It selects minimal dependencies for data access, visualization, and browser validation, then records installation and execution details. Keeping the environment setup as an explicit stage makes failures easier to attribute: a broken run can be traced to data access, dependency configuration, rendering code, or interaction logic.

\paragraph{Main VIS Design Loop.}
The main design loop consists of three roles: Planner, VIS Designer, and Evaluator. The Planner translates the task and EDA Report into a concrete specification covering visual encodings, coordinated panels, interaction state, risks, and validation checks. The VIS Designer implements or repairs complex views, such as 3D volume rendering, progressive streaming, custom brushing, and linked selections. The Evaluator tests the live VIS App against the task goals and returns a \emph{Feedback, Fixes} report. The Orchestrator then incorporates these findings and repeats the loop until no blocking issues remain.

\paragraph{Custom Skills and MCP Connectors.}
The harness extends the base code agent with custom skills and MCP Connectors. In our implementation, the {\em VIS Skill} provides SciVis design conventions; the {\em Frontend Skill} guides library selection and interaction implementation; {\em Playwright-MCP} \cite{playwright_mcp_docs} is crucial for browser inspection, screenshots, console checks, and interaction tests; the {\em Scratchbook Skill} supports durable lesson capture; and a {\em Memory Retrieval Skill} is available for future memory-enabled runs. These tools are part of the method because they determine what the agents can inspect and verify, not only what they can generate as code.

\paragraph{Memory Maintainer.}
After a run, the Memory Maintainer distills insights from the scratchbook and artifacts into a hierarchical, human-readable wiki inspired by LLM wiki memory~\cite{karpathy2024llmwiki}. We use markdown pages rather than embedding-only retrieval so that agents can browse, cite, and audit prior lessons. We leave further ablation for future work to ultimately determine the exact impact.

\subsection{Artifacts and Evaluation}

Agents communicate through explicit artifacts rather than implicit conversation history. The EDA agent writes the \emph{EDA Report}; the Planner writes detailed \emph{Design and Implementation Report}; the Environment Builder builds the Coding Environment; the Orchestrator and VIS Designer produce the VIS App; the Evaluator analyzes this VIS App and provides \emph{Feedback} and \emph{Suggested Fixes}; and the Memory Maintainer retrieves \emph{Previous Insights} and then records new ones, entered by the subagents in the scratchbook. These artifacts not only provide contracts between stages, but also make the reasoning traceable to the human user.

Layered validation and evaluation are essential components for the closed-loop optimization. 
During implementation, Playwright-MCP loads the live application, exercises controls, checks linked-view synchronization, captures screenshots, and reports console errors. These checks detect mechanistic failures, including blank panels, incorrect color mappings, broken brushing, and layout overflow. 
However, the visualization evaluation also requires the agent to perceive the overall system in action: it must judge whether a series of interactions with different visualization elements address the original analysis tasks.  
This makes the validation problem substantially more complex than in many other coding tasks, such as ordinary website/application design.
Beyond mechanistic validation, the Evaluator therefore also scores the system as a whole on whether the original tasks can be achieved from the displayed visualization. The findings then feed back into the main VIS design loop, where the Orchestrator either fixes the issue directly or routes it to the VIS Designer. The final output is the validated VIS App together with the artifact trail used to create and assess it.

\section{Results and Discussion}
\label{sec:results}

We evaluated the VIS co-scientist on the IEEE SciVis Contest tasks using only a high-level instruction: ``solve the tasks in this SciVis contest folder and provide a visualization solution''. These contests are useful stress tests because they combine scientific data, underspecified goals, and task-driven validation rather than isolated chart generation. We focus first on the 2025 contest because it is the most recent one with selected winners, enabling comparison against human contest outcomes.

\subsection{Case Study: SciVis Contest on Materials Discovery}

The 2025 SciVis Contest asks participants to support the discovery of recycled aluminum alloy candidates from a high-dimensional simulation dataset. The tasks require a global design-space overview, explanations of composition--phase--property relationships, and interactive steering for multi-objective optimization. A baseline coding agent, with the same prompt, produced standalone embeddings and correlation plots that partially address the tasks; these were useful analysis artifacts but lacked a cohesive, detailed, interactive visualization design to complete the stated tasks. Moreover, in some cases, they fail to follow key instructions due to a lack of a sophisticated validation harness (examples are provided in supplementary material). 

\begin{figure}[th]
\centering
\includegraphics[width=\linewidth]{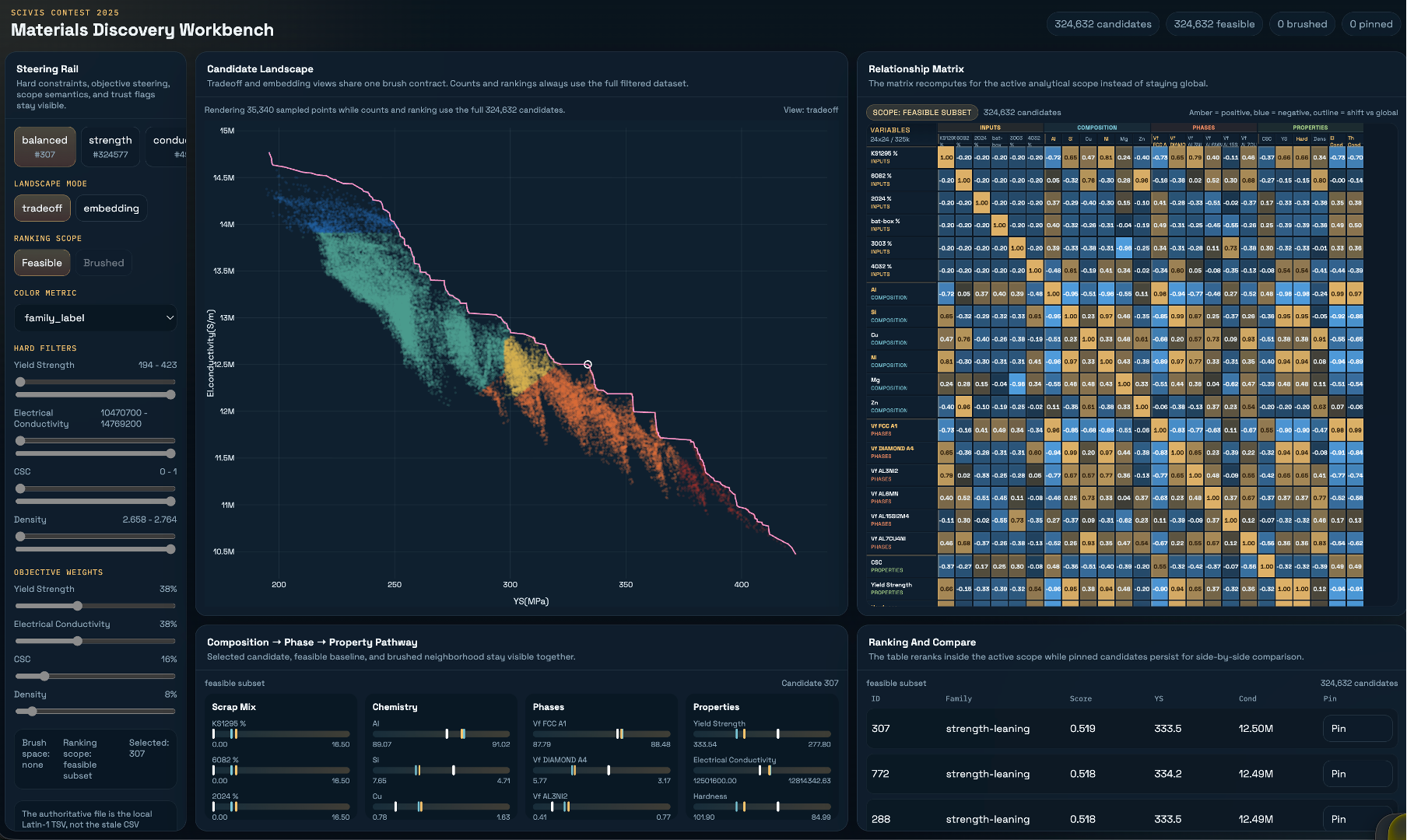}
\caption{VIS co-scientist output for the 2025 materials-discovery case study: a candidate landscape, embedding view, relationship matrix, pathway panel, and ranking/compare view linked in one workspace.}
\label{fig:scivis-contest-2025}
\end{figure}

We ran VIS co-scientist using Codex with GPT-5.4 as the coding model. The run for this use case took approximately two hours and used 368,616 input tokens, an additional 5,183,488 cached tokens, and 58,374 output tokens, including 19,964 reasoning tokens. The generated VIS App (Figure~\ref{fig:scivis-contest-2025}) used a candidate landscape, a PCA embedding mode, a scope-aware relationship matrix, a composition--phase--property pathway panel, and a ranking/compare area. These views were linked through a shared filter with brushing interactions. 
A supplementary video demonstrate the generated VIS app can be viewed at: \url{https://youtu.be/kAuXA_L_17c}.
In the exploratory data analysis report, the agent found that the authoritative source was not the stale CSV path described in the metadata, but a tab-delimited file containing 324,632 candidate alloys and 70 named variables, after padded blank columns were removed. It also detected fully empty variables, created a stable synthetic candidate identifier, and selected \texttt{YS(MPa)} versus \texttt{El.conductivity(S/m)} as the primary trade-off landscape, as those variables exhibited the dominant contest-relevant relationship.

The application recovered several domain-relevant findings. It organized the design space around a strength-conductivity Pareto front while the ideal candidates are selected: balanced candidate is 307, strength-biased candidate shifted to 324577 and the conductivity-biased candidate shifted to 496, which are all visualized in the candidate landscape. 
The relationship matrix and pathway views suggested a composition--phase--property explanation: strength rises with higher Si, Ni, and \texttt{Vf\_DIAMOND\_A4}, while conductivity rises with Al and \texttt{Vf\_FCC\_A1}. 

The final Evaluator marked the application as passing all criteria. 
It also verified that brushing in the tradeoff view changed the relationship matrix, pathway, and ranking scope to the brushed neighborhood, and that switching to embedding mode cleared that local scope. The main residual issues were readability and speed: the relationship matrix remained dense, brushing larger regions take several seconds.

\subsection{Structured Expert Audit}

We also conducted a structured audit with five author-team visualization researchers (see Table~\ref{tab:human-eval}). This audit served as a bounded validity check: whether the expected artifacts are meaningful, whether the application was mechanically usable, whether generated views supported the stated contest tasks, and whether curated domain insights (gathered from winning entries' reports) could be recovered from the tool. Five participants reviewed the EDA report, planning and evaluation artifacts, the final VIS App, and task-specific domain statements derived from the contest questions and materials \cite{ai2026scivisagentbench}. The questionnaire is included in the supplemental material.

\begin{table}[th]
\centering
\fontsize{6.5}{8.5}\selectfont
\caption{Structured expert-audit summary for the 2025 materials-discovery case study. Scores use a 1--5 Likert scale (1 = unacceptable, 5 = excellent) with five visualization researchers.}
\label{tab:human-eval}
\setlength{\tabcolsep}{3pt}
\begin{tabular}{p{0.80\linewidth}cc}
\hline
Question & Mean & Range \\
\hline
EDA report makes correct assumptions about the data and tasks & 4.6 & 4--5 \\
EDA identifies key implementation insights and constraints & 4.6 & 4--5 \\
Proposed design follows established visualization principles & 4.0 & 3--5 \\
Evaluation documents give logical next-iteration guidance & 4.4 & 4--5 \\
Visual encodings are appropriate for the data and intent & 3.8 & 2--5 \\
Interface is visible and responsive for exploration & 4.0 & 3--5 \\
Coordinated views support sensemaking & 4.6 & 4--5 \\
Dashboard enables stated contest tasks & 4.2 & 3--5 \\
Removing all-NaN fields was correct & 5.0 & 5--5 \\
\texttt{KS1295[\%]} improves strength but hurts thermal/electrical performance & 3.4 & 2--5 \\
Strength and thermal/electrical performance form a fundamental tradeoff & 5.0 & 5--5 \\
Hardness and yield strength are strongly correlated & 5.0 & 5--5 \\
\texttt{6082[\%]} strongly influences Cu, Ti, and Zn & 2.4 & 1--4 \\
\hline
\end{tabular}
\end{table}

The audit separates task completion from the visualization design quality. Experts gave high scores to EDA assumptions, implementation insights, coordinated views, and task completion; open-ended comments identified brushing, filter propagation, candidate ranking, and the task-specific application as strengths. Lower or more variable scores concerned visual encoding and interface polish: missing legends, ambiguous color semantics, dense panel layout, limited control over available variables, and reduced responsiveness after brushing. The application also lacked the visual-encoding novelty that characterized the winners. We interpret this as a difference in optimization target rather than a weakness: the harness aimed to complete the specified tasks with established encodings, while contest-winning solutions often call for a distinctive visualization contribution.

The domain-insight questions show the same split. Experts unanimously agreed with removing all-NaN fields, the strength-performance tradeoff, and the hardness-yield-strength relationship. Agreement was weaker for the \texttt{KS1295[\%]} statement and low for the \texttt{6082[\%]} influence statement, suggesting that automatically surfaced correlations need clearer provenance.
The case highlights a difference between autonomous generation by the VIS co-scientist and human visualization research. Overall, the agent performed well despite vague contest-level instructions, but lacked nuanced prior knowledge of a visualization expert.
The experts still considered the system coherent and technically advanced, but the weaknesses show why useful task-oriented prototypes are not equivalent to human-led contest entries.

\subsection{Other SciVis Contests}

We also applied the agent harness to available SciVis Contests from 2021, 2023, 2024, and 2026 (refer to Figure~\ref{fig:teaser} and the supplemental material). All runs produced mechanically valid VIS Apps, suggesting that the approach transfers across scientific domains, even from a high-level contest instruction. These cases covered different file formats, scientific questions, and interaction demands, indicating that the harness was not tuned only to the materials case. Results were strongest when tasks were decomposed into data profiling, established encodings, coordinated views, and task-specific interactions: the agent found data-quality issues, built custom applications rather than generic plots, linked brushing and filtering, surfaced rankings or comparable outputs, and used evaluation feedback to fix implementation failures.

The recurring limit was a complex dynamic, multi-step visual reasoning, especially in 3D views. Screenshots and DOM checks can show that an interface is rendered, but they weakly test intermittent sequences of camera movement, filtering, and linked state changes; the agent, therefore, cannot yet reliably critique temporal visual behavior, which is often central to visual representation design and design studies. When a task depended on volumetric exploration or continuous camera interaction, evaluation could catch blank renders and obvious state failures but had limited leverage over whether the interaction sequence supported scientific reasoning. This limitation is not a property of our harness but a broader limitation of current LLM-based agentic systems. As in the audit, the outputs were useful task-oriented prototypes, but lacked human-level novelty and human-level perception of the visual interface. We plan to submit a VIS co-scientist-created entry to the 2026 SciVis Contest as a stronger external evaluation.

\subsection{Toward Memory and Self-Improving VIS Co-Scientists}

While current contest results establish a high-performing baseline, that does not rely on memory from the wiki, we anticipate that this infrastructure enables a transition from isolated task execution to self-improving co-scientists that start the task by retrieving previously successful design patterns and data-access strategies and avoid already encountered pitfalls. We envision this as an essential step toward reducing token costs and ensuring the agents to improve over time. Future evaluation will quantify the performance delta of memory-augmented agents to determine how persistent knowledge capture fundamentally alters the capability of the VIS co-scientist.


\section{Conclusions and Future Work}


This work demonstrates that end-to-end VIS App design is becoming feasible for agentic AI systems using the proposed harness. 
Given only a dataset and high-level tasks, a VIS co-scientist can inspect data, plan, analyze, implement views, validate interactions, and deliver VIS Apps for domain scientists. 
As such AI systems improve, the VIS community’s role can potentially shifts upward: defining abstractions for scientific sensemaking, formalizing evaluation criteria, encoding perceptual knowledge into agent workflows, and judging when a visualization makes a scientific claim trustworthy rather than merely interactive. 
In this sense, the VIS co-scientist is a necessary part of the broader AI co-scientist vision, because autonomous science needs not only on computation, but also exploratory interfaces and interpretable visual evidence.

At the same time, our results expose the boundary of current models. 
They can repair interfaces and compose established visual encodings, but still lack the visual perception and temporal understanding needed for closed-loop design of dynamic and novel interfaces.
The implication is not to abandon application building, but to redefine it: future VIS research should make visualization knowledge, perceptual judgment, human goals, and evaluation protocols first-class components of AI scientific systems.
Another important shortcoming of current system comes from assessing uncertainty and ensuring trustworthiness through the long-task duration.
In a sense, the frontier of research is shifting from designing a single artifact/solution to designing the robust process that creates, critiques, and validates such artifacts with domain scientists.

\acknowledgments{
This research was supported in part by the U.S.\ National Science Foundation through grants IIS-2101696, OAC-2104158, and IIS-2401144, and the U.S.\ Department of Energy through grant DE-SC0023145. 
This work was also performed under the auspices of the U.S. Department of Energy by Lawrence Livermore National Laboratory under Contract DE-AC52-07NA27344. The work is partially funded by DOE ECRP 51917/SCW1885. The manuscript is reviewed and released under LLNL-CONF-2018939.
}

\bibliographystyle{abbrv-doi}

\vspace{-0.05in}
\bibliography{refs}

\clearpage
\setcounter{section}{0}
\setcounter{figure}{0}
\setcounter{table}{0}
\setcounter{page}{1}
\onecolumn
\section{Appendix: Evaluation Setup}
\label{sec:supplemental_materials}

\subsection{Evaluation of Individual Agent Reasoning Artifacts}
As discussed in the Method section of the main paper, the agent generates intermediate artifacts that describe the subagents' completed tasks, including exploratory data analysis, planning, and evaluation. 
They provide insights into key decisions throughout the design and implementation process, helping reveal the agent's behavior and inner workings. The first two focus on the generated EDA report, and the next two focus on the VIS App design plan, individually, and the following three focus on 

\begin{myitemize}
\vspace{-0.05in}

    \item The EDA report makes a correct assumption about the data and tasks.
    
    \item The EDA report identifies key insights for implementing the overall VIS App, e.g., the dataset's structure, data quality issues, implementation constraints, etc.
    
    \item The proposed design follows established VIS principles to produce meaningful visual encodings, or justifies a deviation from the norm. 
    
    \item The evaluation documents provide a detailed and logical argument of what is not working and what needs to be improved in the next iteration.
    

\vspace{-0.05in}
\end{myitemize}

\subsection{Appendix: Evaluation of VIS App Quality}
In addition to the intermediate steps, the most essential evaluation comes from the final VIS App produced by the multi-agent system:

\begin{myitemize}
\vspace{-0.05in}

    \item The VIS App's chosen visual encodings are appropriate for the data and analytical intent. 
    
    \item The interface and individual visualizations are visible without artifacts. Moreover, the interactions enable real-time exploration.

    \item The coordinated views work together to support sense making, e.g., selections and filters propagate across linked views. 
    
    \item The dashboard enables the stated task(s) in the contest to be completed effectively, e.g., by understanding property correlations and enabling targeted optimization of material properties.
    
\vspace{-0.05in}
\end{myitemize}

\subsection{Evaluation of Extracted Insight from the Tool}
Does the VIS App or the exploratory data analysis report allow the user to uncover insights about the data? We extracted and collected a few statements from the 2025 SciVis Contest-winning publications to provide a baseline assessment:

\begin{myitemize}
\vspace{-0.05in}

\item  Missing value: the dimensions Vf MG2ZN3, T MG2ZN3, and T AL3X were removed due to containing only NaN value.
\item  Increasing KS1295[\%] improves YS (good) but hurts overall performance (Thermal conductivity, Thermal diffusivity, Electrical conductivity, Heat capacity, Thermal expansion)
\item There is a fundamental trade-off between strength and thermal performance
\item  Hardness (Vickers) and YS (MPa) are strongly correlated.
\item 6082[\%] strongly influences Cu, Ti, Zn

\vspace{-0.05in}
\end{myitemize}

For each question, rate on a scale of 1-5:
\begin{myitemize}
\vspace{-0.05in}

    \item \textbf{5 (Excellent):} Exceeds expectations; demonstrates expert-level quality
    \item \textbf{4 (Good):} Meets expectations with minor issues or room for improvement
    \item \textbf{3 (Adequate):} Acceptable but has noticeable limitations or gaps
    \item \textbf{2 (Poor):} Significant deficiencies that impact usability or correctness
    \item \textbf{1 (Unacceptable):} Fails to meet basic requirements; unusable or incorrect
\vspace{-0.05in}
\end{myitemize}

\clearpage

\section{Appendix: SciVis Contest VIS App Screenshots}
\label{sec:supplemental-scivis-contest-screenshots}
\begin{figure*}[hbtp]
\centering
\includegraphics[width=\textwidth]{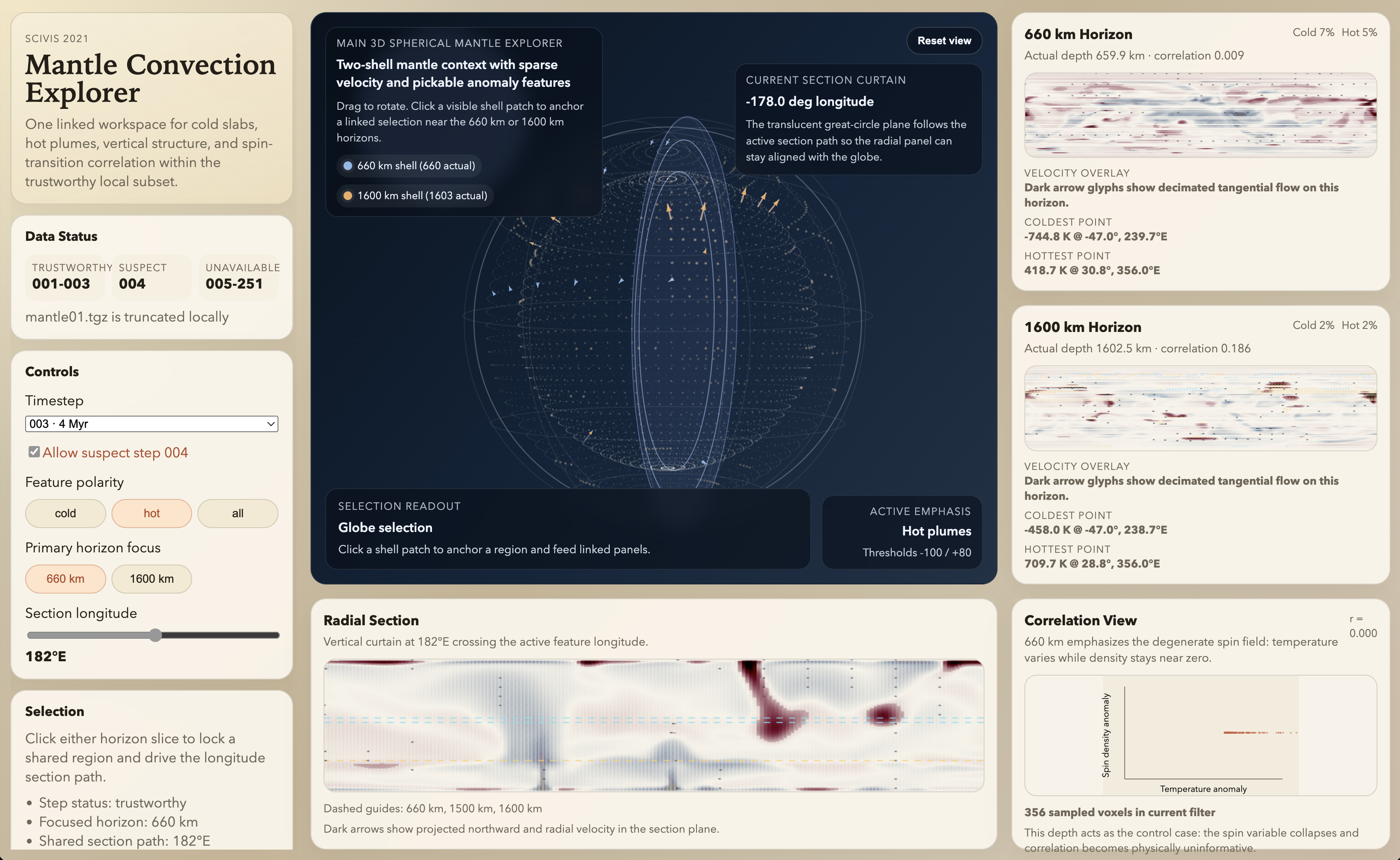}
\vspace{-2mm}
\caption{VIS App screenshot for the 2021 SciVis Contest case.}
\label{fig:supp-scivis-contest-2021}
\end{figure*}

\begin{figure*}[hbtp]
\centering
\includegraphics[width=\textwidth]{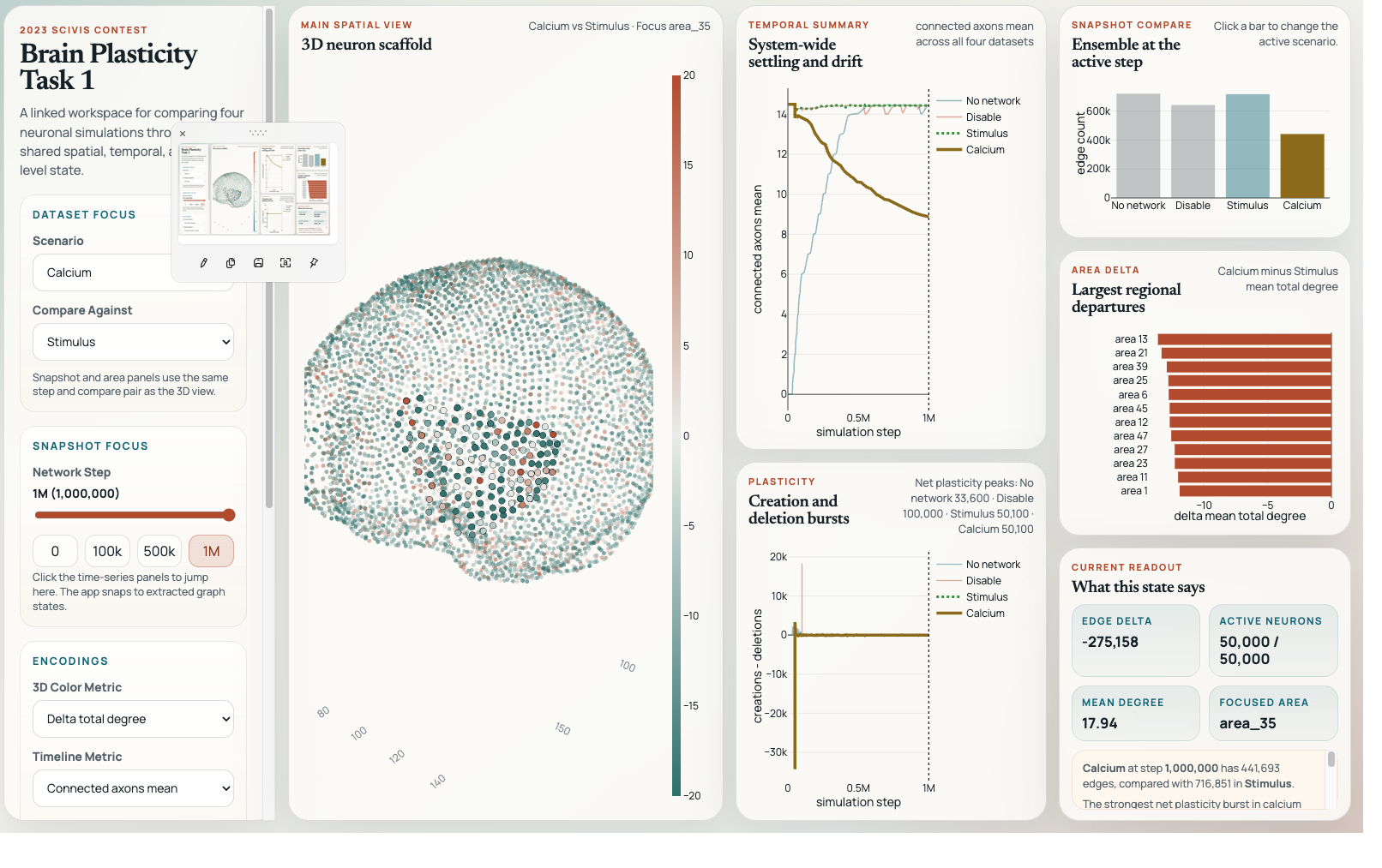}
\vspace{-2mm}
\caption{VIS App screenshot for the 2023 SciVis Contest case.}
\label{fig:supp-scivis-contest-2023}
\end{figure*}

\begin{figure*}[hbtp]
\centering
\includegraphics[width=\textwidth]{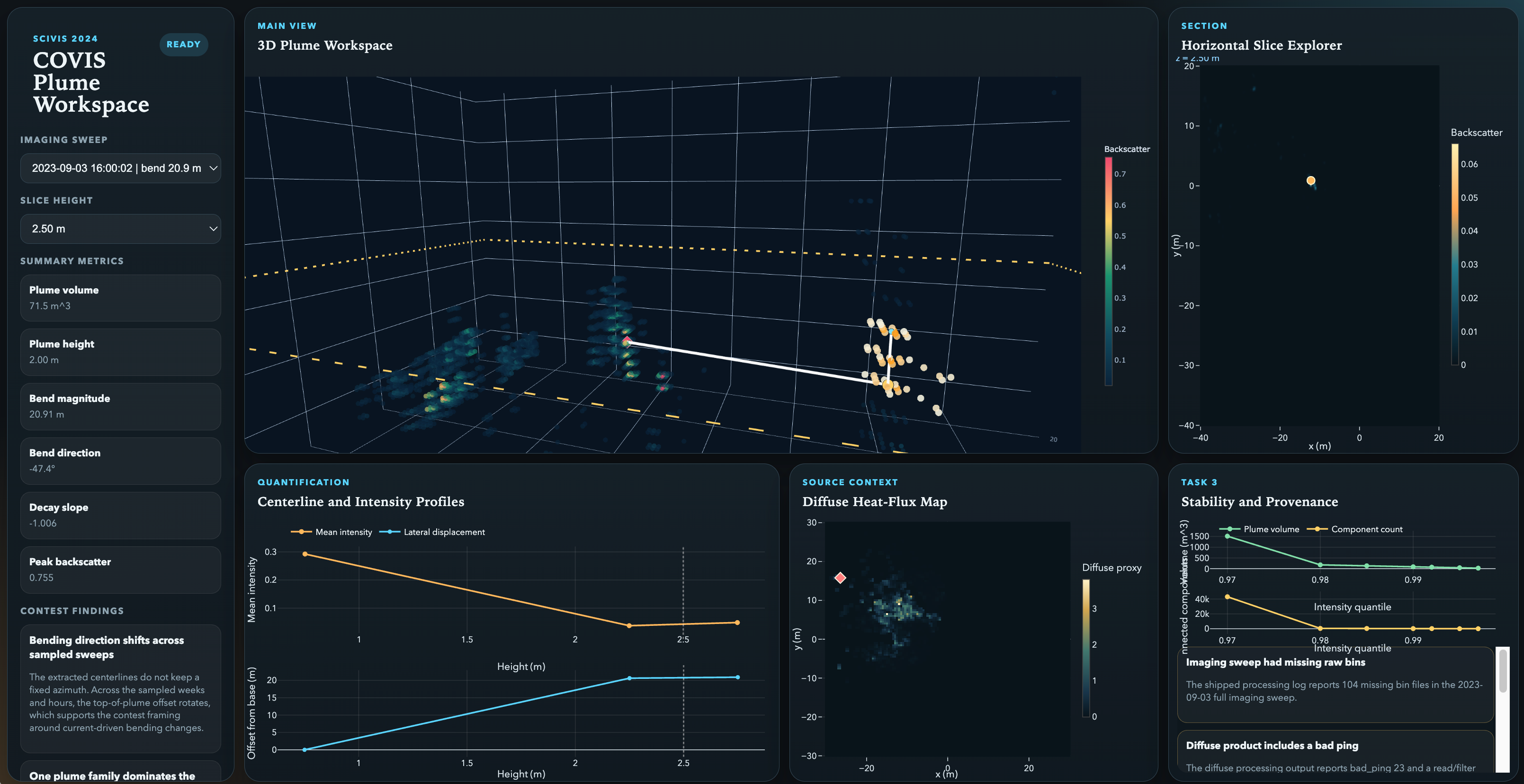}
\vspace{-2mm}
\caption{VIS App screenshot for the 2024 SciVis Contest case.}
\label{fig:supp-scivis-contest-2024}
\end{figure*}

\begin{figure*}[hbtp]
\centering
\includegraphics[width=\textwidth]{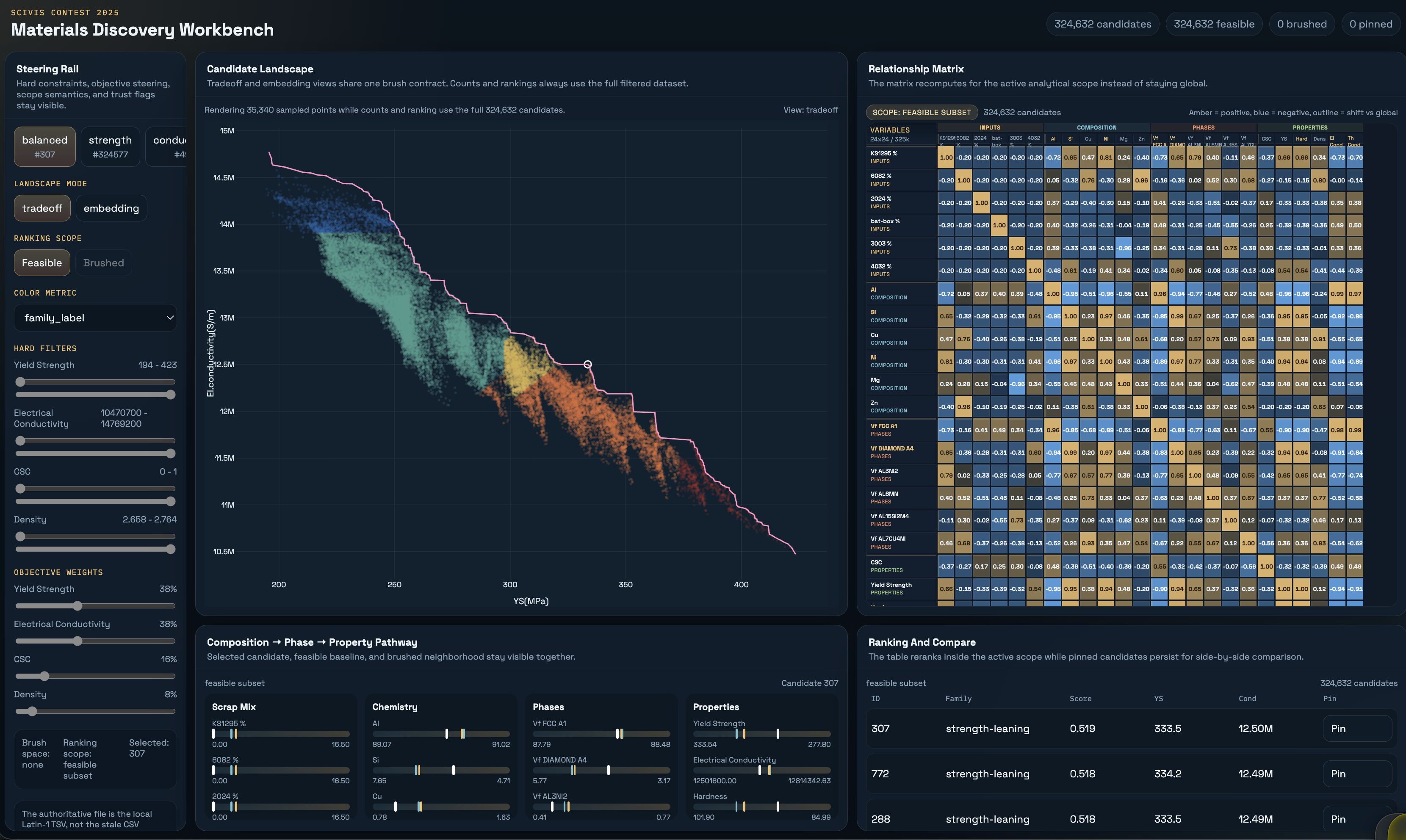}
\vspace{-2mm}
\caption{VIS App screenshot for the 2025 SciVis Contest case.}
\label{fig:supp-scivis-contest-2025}
\end{figure*}

Figures~\ref{fig:supp-scivis-contest-2021}--\ref{fig:supp-scivis-contest-2025} show full-width screenshots of VIS Apps generated for SciVis Contest cases across multiple years.

\clearpage

\section{Appendix: Baseline 2025 Coding-Agent Output}
\label{sec:supplemental-baseline-2025}

\begin{figure*}[hbtp]
\centering
\includegraphics[width=0.48\linewidth]{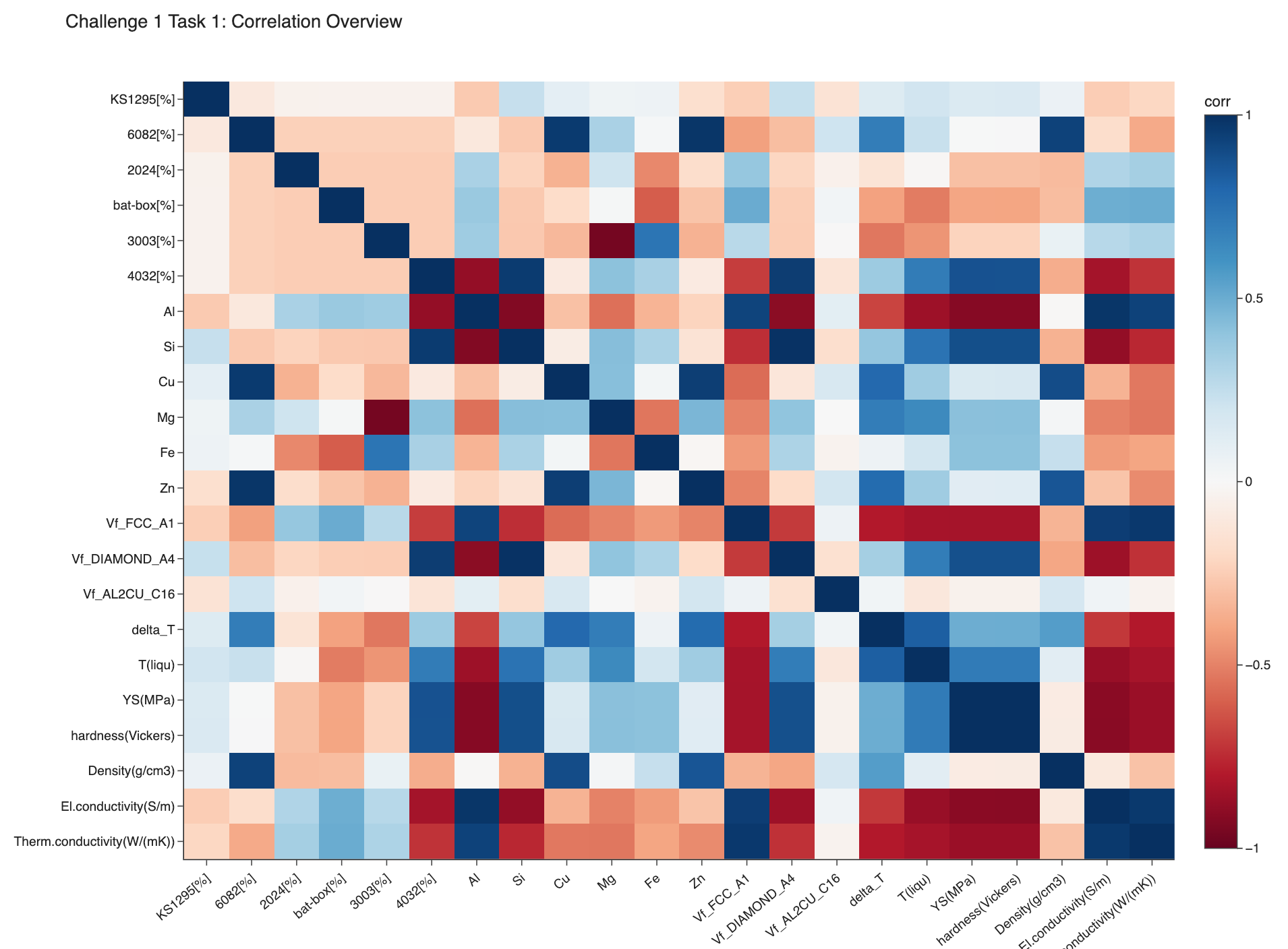}
\hfill
\includegraphics[width=0.48\linewidth]{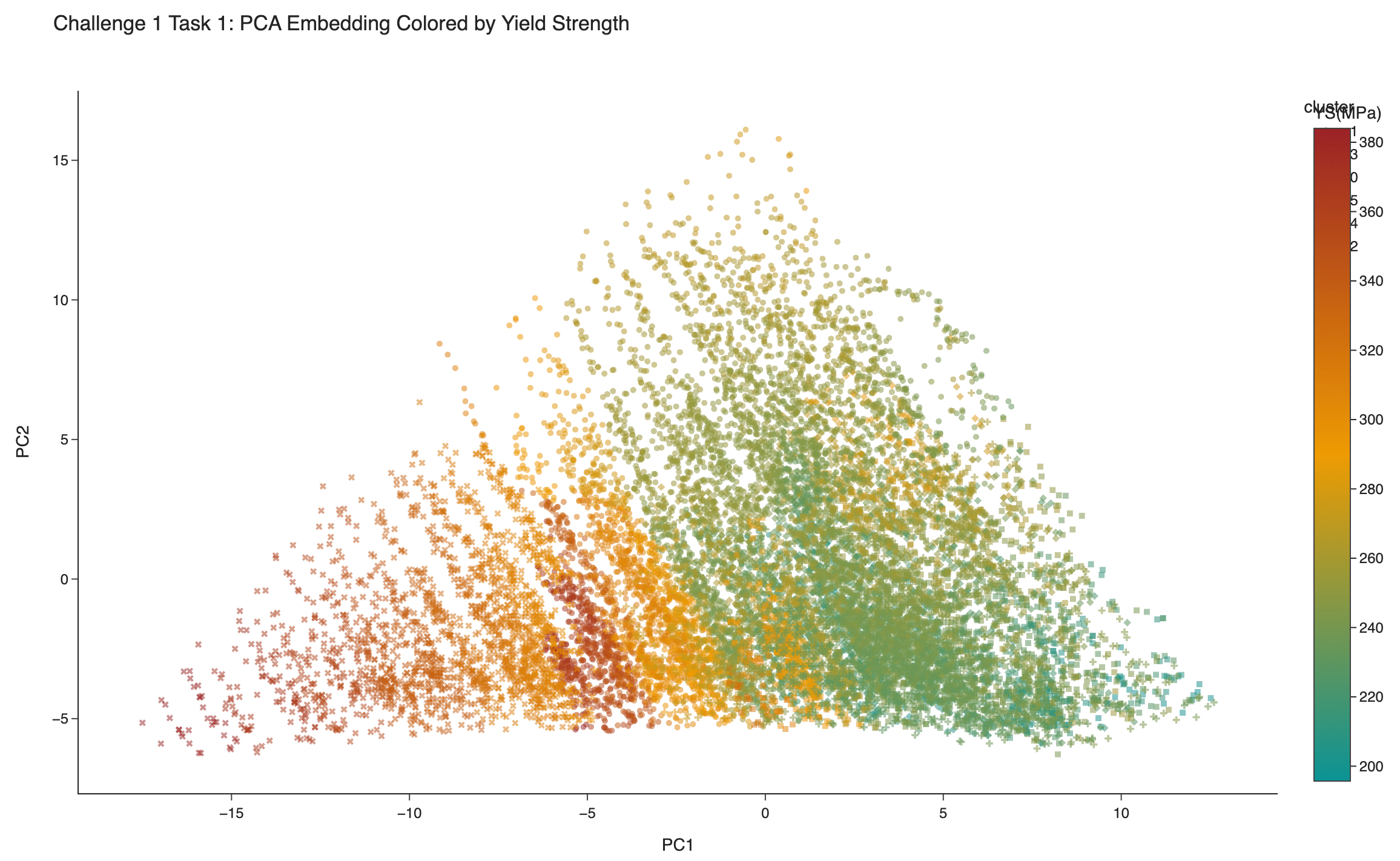}

\vspace{2mm}
\includegraphics[width=0.48\linewidth]{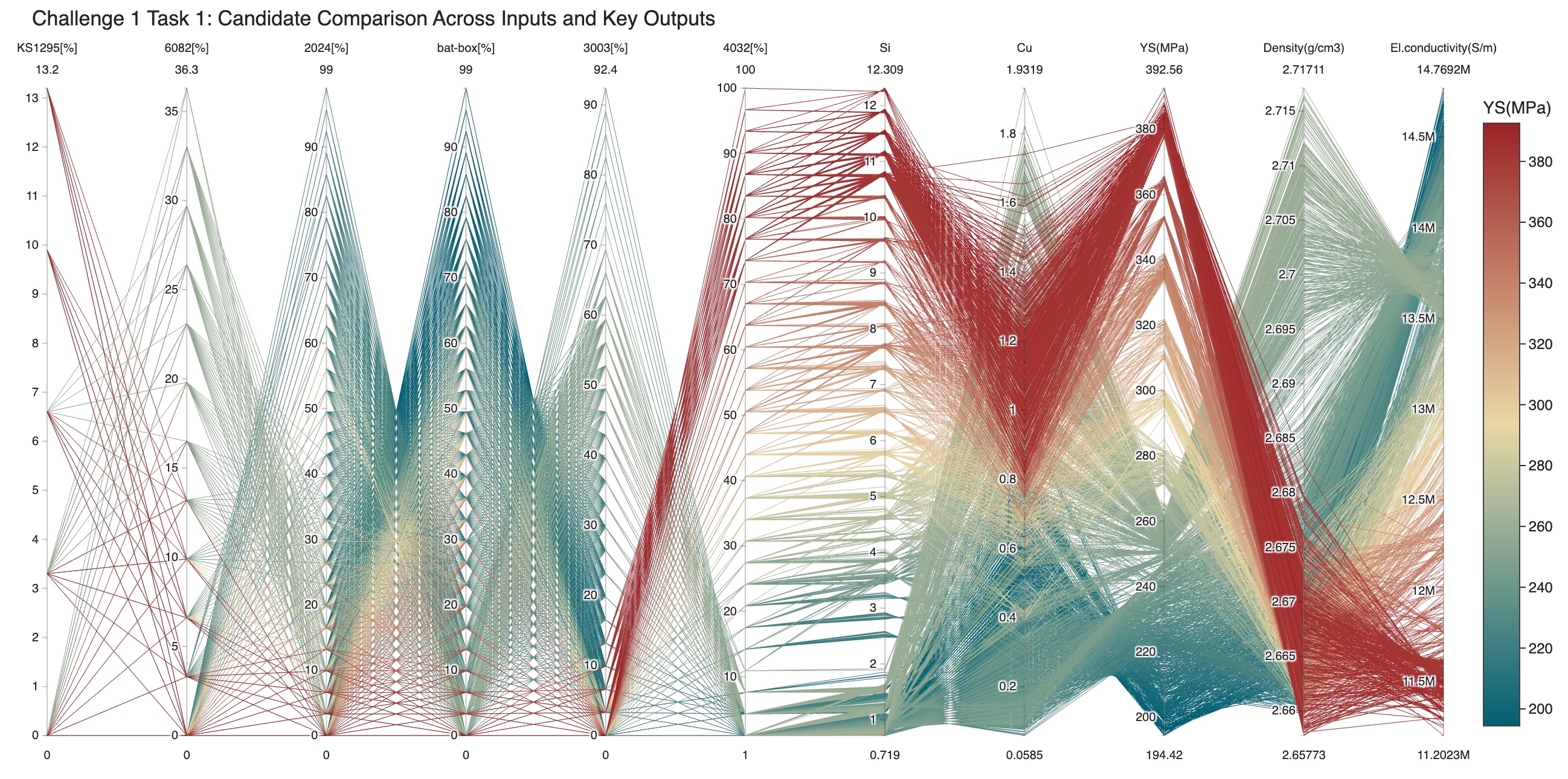}
\hfill
\includegraphics[width=0.48\linewidth]{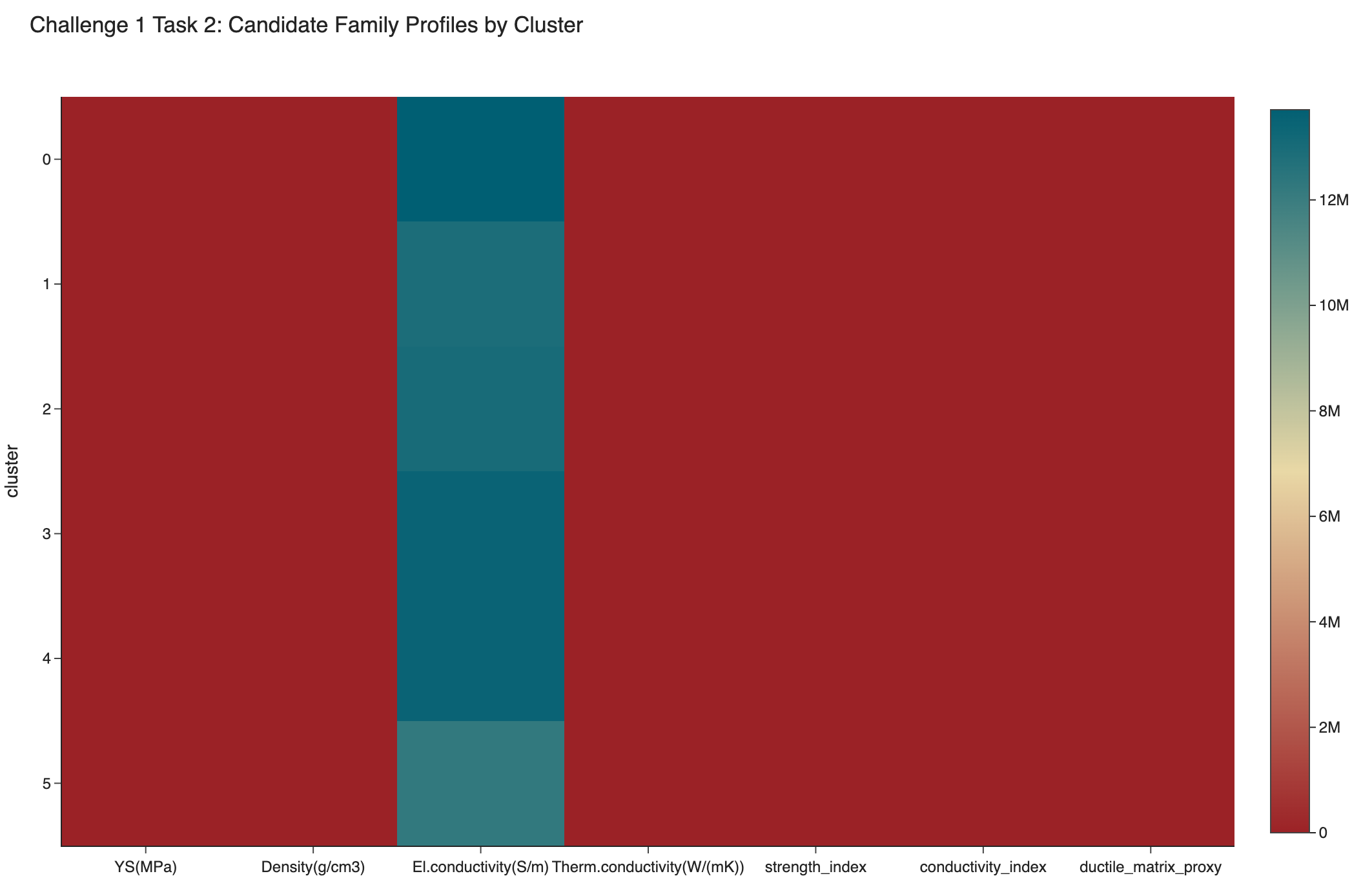}
\vspace{-2mm}
\caption{Baseline coding-agent plots for the 2025 SciVis Contest materials-discovery case. Top left: correlation overview for Challenge 1, Task 1. Top right: PCA embedding for Challenge 1, Task 1. Bottom left: parallel-coordinates view for Challenge 1, Task 1. Bottom right: candidate-family summary for Challenge 1, Task 2.}
\label{fig:supp-baseline-challenge1-task1}
\end{figure*}

\begin{figure*}[hbtp]
\centering
\includegraphics[width=0.48\linewidth]{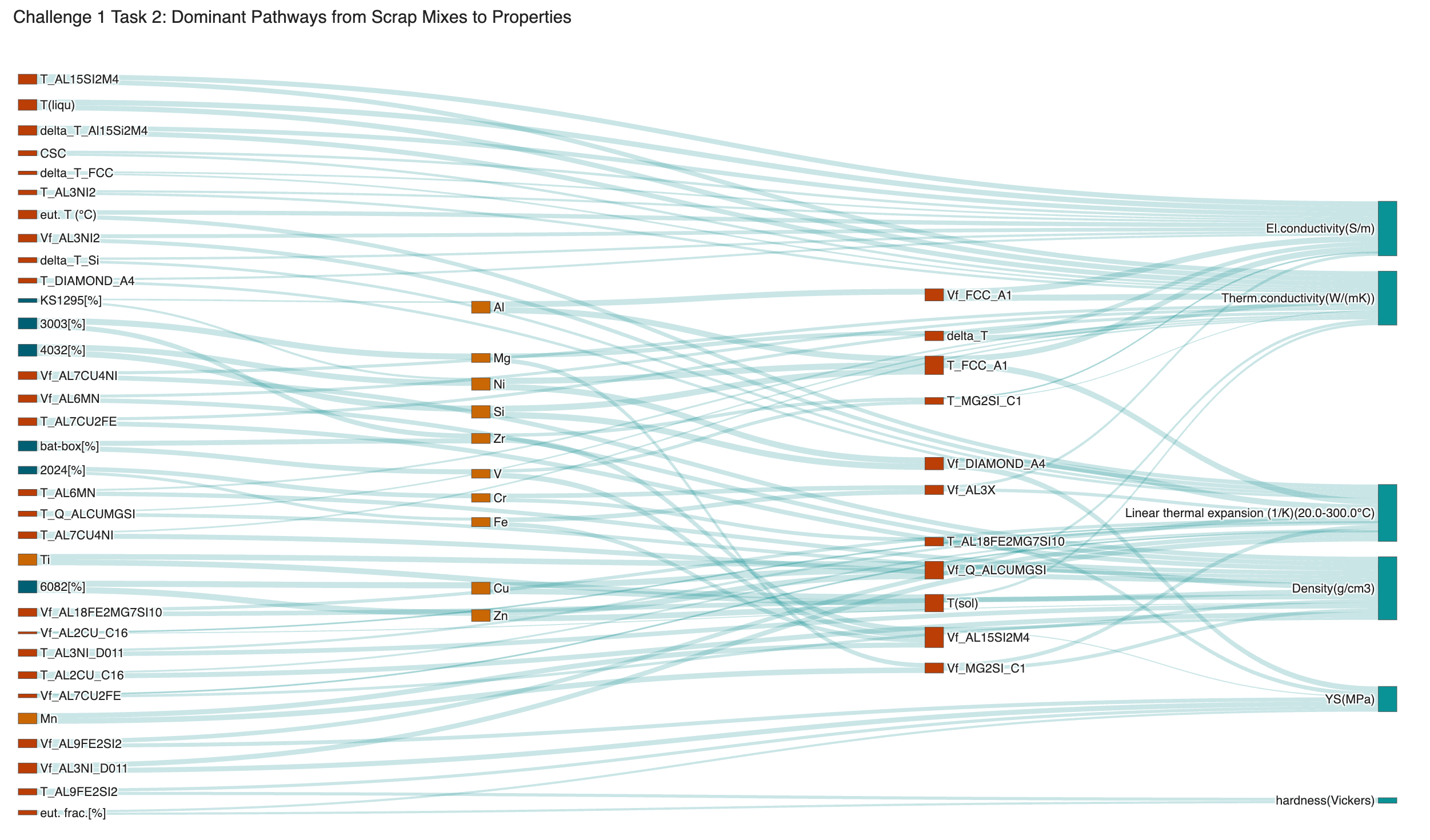}
\hfill
\includegraphics[width=0.48\linewidth]{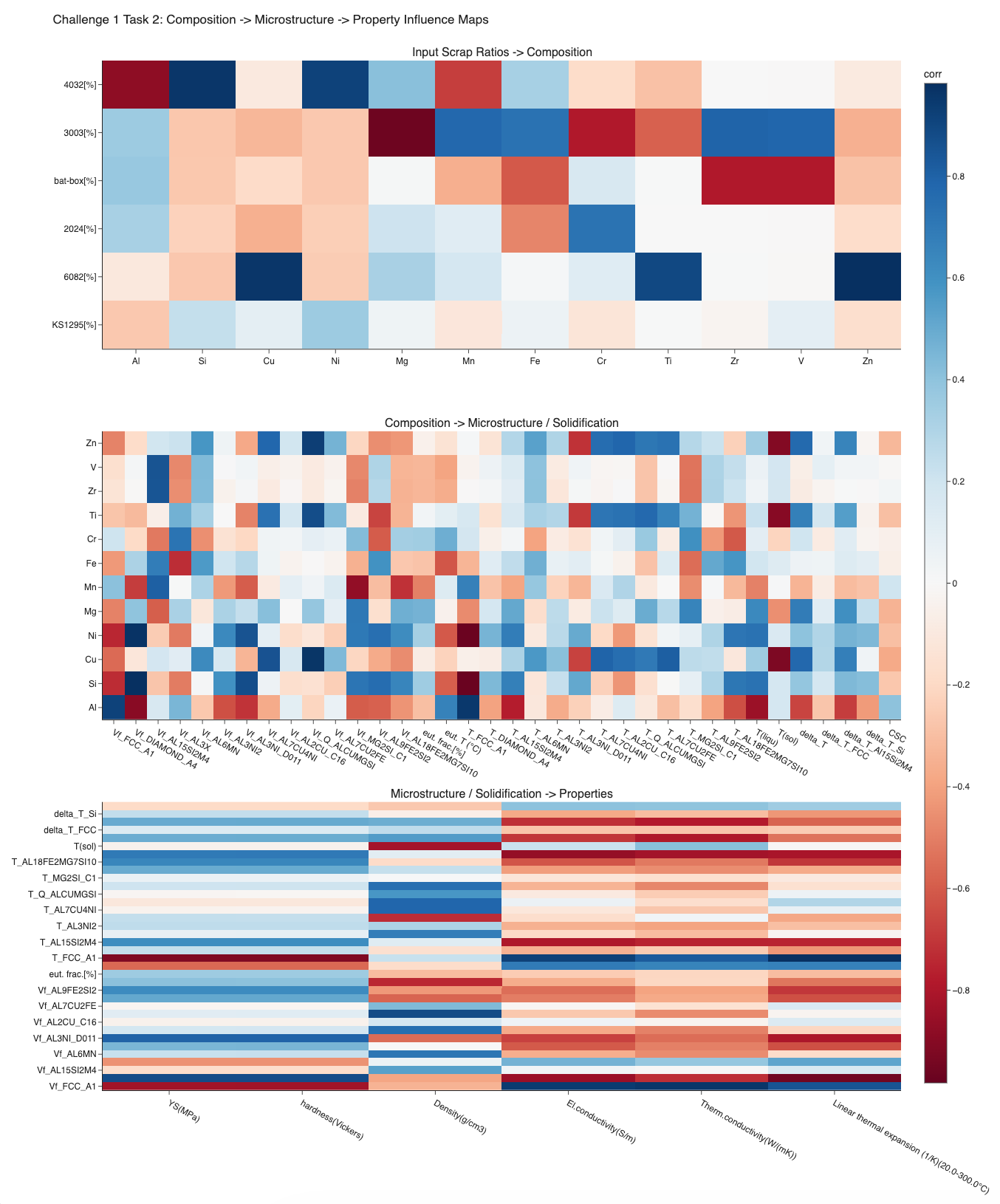}

\vspace{2mm}
\includegraphics[width=0.48\linewidth]{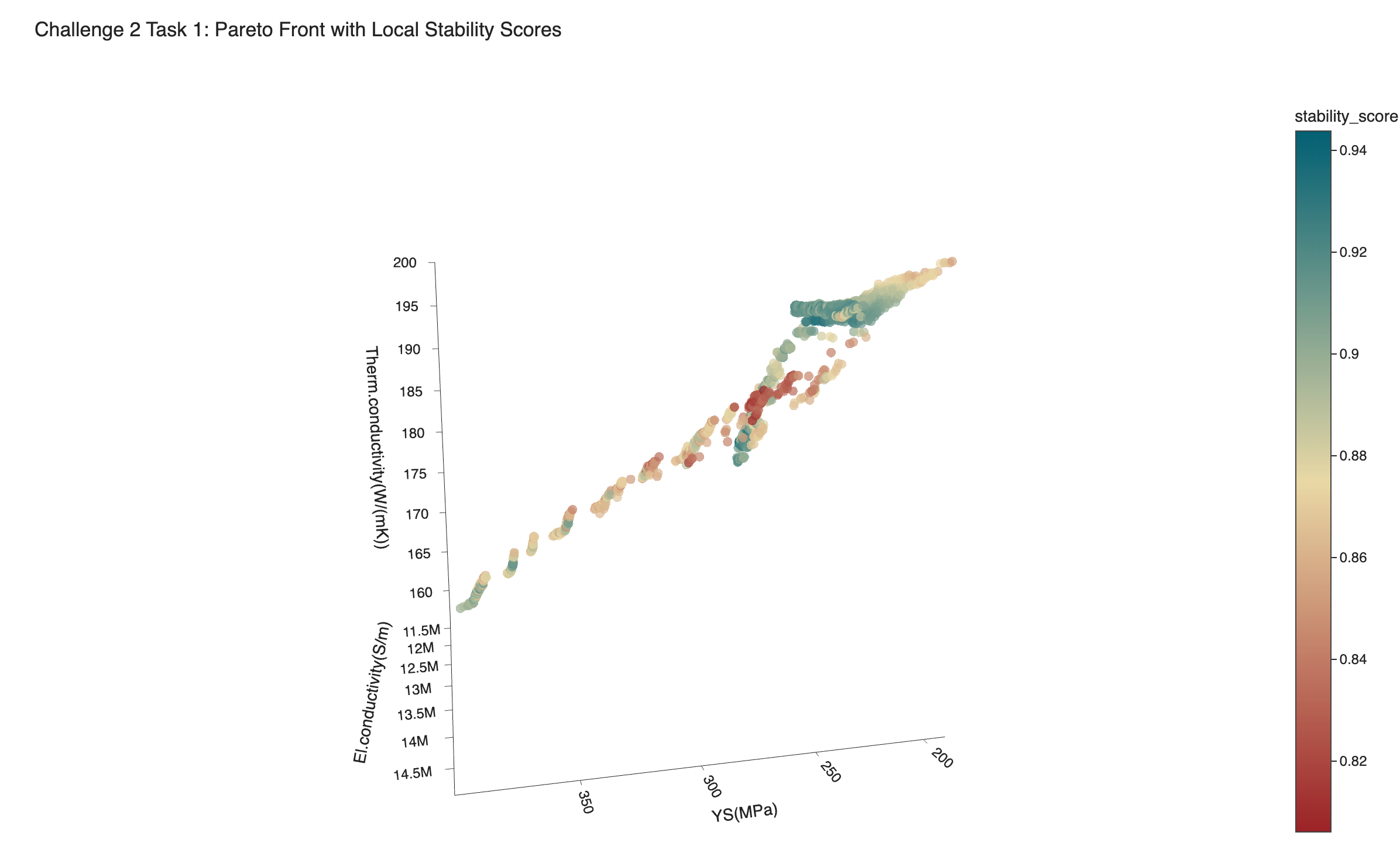}
\hfill
\includegraphics[width=0.48\linewidth]{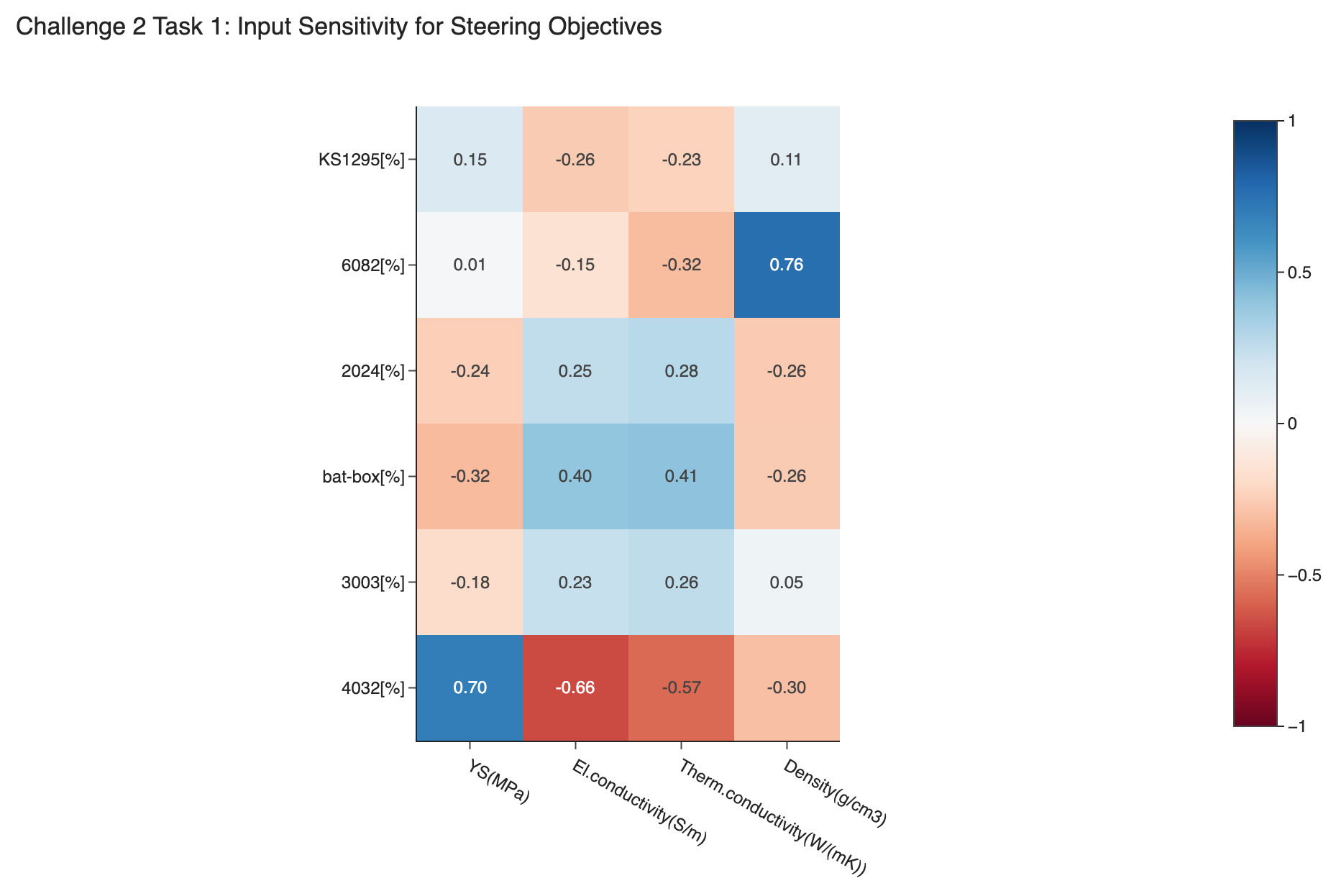}
\vspace{-2mm}
\caption{Additional baseline coding-agent plots for the 2025 SciVis Contest materials-discovery case. Top left: composition--phase--property pathway view for Challenge 1, Task 2. Top right: stage-influence summary for Challenge 1, Task 2. Bottom left: Pareto-front plot for Challenge 2, Task 1. Bottom right: sensitivity heatmap for Challenge 2, Task 1.}
\label{fig:supp-baseline-challenge1-task2-challenge2-task1}
\end{figure*}

Figure~\ref{fig:supp-baseline-challenge1-task1} and Figure~\ref{fig:supp-baseline-challenge1-task2-challenge2-task1} show the plotting-based solution produced by the baseline coding agent for the 2025 SciVis Contest when it was instructed directly, without the VIS co-scientist agent harness. These outputs provide useful analysis views, but they are standalone plots rather than a validated, coordinated VIS App.

\end{document}